\documentclass{article}
\usepackage{spconf,amsmath,graphicx}
\usepackage{multirow}

\title{RNN-TRANSDUCER WITH LANGUAGE BIAS FOR END-TO-END MANDARIN-ENGLISH CODE-SWITCHING SPEECH RECOGNITION}
\name{Shuai Zhang$^{1,2}$, Jiangyan Yi$^{1}$, Zhengkun Tian$^{1,2}$, Jianhua Tao$^{1,2,3}$, Ye Bai$^{1,2}$}

\address{$^1$NLPR, Institute of Automation, Chinese Academy of Sciences, China\\
	$^2$School of Artificial Intelligence, University of Chinese Academy of Sciences, China \\
	$^3$CAS Center for Excellence in Brain Science and Intelligence Technology, China}
\begin{document}
%
\maketitle
\begin{abstract}
Recently, language identity information has been utilized to improve the performance of end-to-end code-switching (CS) speech recognition. However, previous works use an additional language identification (LID) model as an auxiliary module, which causes the system complex. In this work, we propose an improved recurrent neural network transducer (RNN-T) model with language bias to alleviate the problem. We use the language identities to bias the model to predict the CS points. This promotes the model to learn the language identity information directly from transcription, and no additional LID model is needed. We evaluate the approach on a Mandarin-English CS corpus SEAME. Compared to our RNN-T baseline, the proposed method can achieve 16.2\% and 12.9\% relative error reduction on two test sets, respectively. 
\end{abstract}

\section{Introduction}
\label{sec:intro}

Code-switching (CS) speech is defined as the alternation of languages in an utterance, it is a pervasive communicative phenomenon in multilingual communities. Therefore, developing a CS speech recognition (CSSR) system is of great interest. 

However, the CS scenario presents challenges to recognition system \cite{ccetinouglu2016challenges}. Some attempts based on DNN-HMM framework have been made to alleviate these problems \cite{li2011asymmetric,guo2018study}. The methods usually contain components including acoustic, language, and lexicon models that are trained with different object separately, which would lead to sub-optimal performance. And the design of complicated lexicon including different languages would consume lots of human efforts.

Therefore, end-to-end framework for CSSR has received increasing attention recently \cite{zeng2018end, Shan2019Investigating, li2019towards}. Examples of such models include connectionist temporal classification (CTC) \cite{graves2006connectionist}, attention-based encoder-decoder models \cite{chan2016listen,bahdanau2016end}, and the recurrent neural network transducer (RNN-T) \cite{graves2013speech,graves2014towards,rao2017exploring,Tian2019Self}. 
\begin{figure}[htb]
	
	\begin{minipage}[b]{1.0\linewidth}
		\centering
		\centerline{\includegraphics[width=8.5cm]{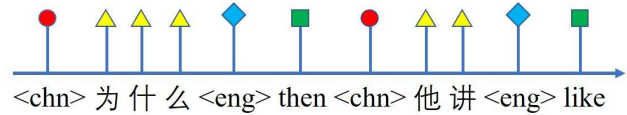}}
		
		\centerline{}\medskip
	\end{minipage}
	\vspace{-35pt}
	\caption{Code-switching distribution diagram.}
	\label{fig:res}
\end{figure}
These methods combine acoustic, language, and lexicon models into a single model with joint training. And the RNN-T and attention-based models trained with large speech corpus perform competitively compared to the state-of-art model in some tasks \cite{Prabhavalkar2017A}. However, the lack of CS training data poses serious problem to end-to-end methods. To address the problem, language identity information is utilized to improve the performance of recognition \cite{zeng2018end, Shan2019Investigating, li2019towards}. They are usually based on CTC or attention-based encoder-decoder models or the combination of both. However, previous works use an additional language identification (LID) model as an auxiliary module, which causes the system complex.

In this paper, we propose an improved RNN-T model with language bias to alleviate the problem. The model is trained to predict language IDs as well as the subwords. To ensure the model can learn CS information, we add language IDs in the CS point of transcription, as illustrated in Fig. 1. In the figure, we use the arrangements of different geometric icons to represent the CS distribution. Compared with normal text, the tagged data can bias the RNN-T to predict language IDs in CS points. So our method can model the CS distribution directly, no additional LID model is needed. Then we constrain the input word embedding with its corresponding language ID, which is beneficial for model to learn the language identity information from transcription. In the inference process, the predicted language IDs are used to adjust the output posteriors. The experiment results on CS corpus show that our proposed method outperforms the RNN-T baseline (without language bias) significantly. Overall, our best model achieves 16.2\% and 12.9\% relative error reduction on two test sets, respectively. To our best knowledge, this is the first attempt of using the RNN-T model with language bias as an end-to-end CSSR strategy. 

The rest of the paper is organized as follows. In Section 2, we review RNN-T model. In Section 3, we describe the intuition of the proposed model. In Section 4, we present the experimental setups, and in Section 5, we report and discuss the experiment results in detail. Finally, we conclude the paper in Section 6.

\section{review of RNN-T}
\label{sec:pagestyle}

Although CTC has been applied successfully in the context of speech recognition, it assumes that outputs at each step are independent of the previous predictions \cite{graves2006connectionist}. RNN-T is an improved model based on CTC, it augments with a prediction network, which is explicitly conditioned on the previous outputs \cite{graves2014towards}, as illustrated in Fig. 2(a). 

Let $\mathnormal{ \mathbf{X} = (\mathbf{x}_{1}, \mathbf{x}_{2}, ... , \mathbf{x}_{T})}$ be the acoustic input sequence, where $T$ is the frame number of sequence. Let $\mathbf{Y} = (\mathnormal{y}_{1}, \mathnormal{y}_{2}, ... , \mathnormal{y}_{U})$ be the corresponding sequence of output targets (without language IDs) over the RNN-T output space $\mathcal{Y}$, and $\mathcal{Y}^{*}$ be the set of all possible sequence over $\mathcal{Y}$. In the context of ASR, the input sequence is much longer than output targets, i.e., $T>U$. Because the frame-level alignments of the target label are unknown, RNN-T augments the output set with an additional symbol, refer to as the $\mathit{blank}$ symbol, denoted as $\phi$, i.e., $\bar{\mathcal{Y}} \in \mathcal{Y} \cup \{\phi\}$. We denote $\hat{\mathbf{Y}} \in \bar{\mathcal{Y}}^{*}$ as an alignment, which are equivalent to $(\mathnormal{y}_{1}, \mathnormal{y}_{2},\mathnormal{y}_{3}) \in \mathcal{Y}^*$ after operation $\mathcal{B}$, such as $\hat{\mathbf{Y}} = (\mathnormal{y}_{1}, \phi, \mathnormal{y}_{2}, \phi, \phi, \mathnormal{y}_{3}) \in \bar{\mathcal{Y}}^{*}$. Given the input sequence $\mathbf{X}$, RNN-T models the conditional probability $P(\mathbf{Y} \in \mathcal{Y}^* | \mathbf{X})$ by marginalizing over all possible alignments:

\begin{equation}
P(\mathbf{Y} \in \mathcal{Y}^* | \mathbf{X})=\sum_{\hat{\mathbf{Y}} \in \mathcal{B}^{-1}(\mathbf{Y})} P(\hat{\mathbf{Y}} | \mathbf{X})
\end{equation}
where $\mathcal{B}$ is the function that removes consecutive identical symbols and then removing any blank from a given alignment in $\bar{\mathcal{Y}}^{*}$.

An RNN-T model consists of three different networks as illustrated in Fig. 2(a). (a) Encoder network (referred to as transcription network) maps the acoustic features into higher level representation $\mathbf{h}_{t}^{enc} = f^{enc}(\{\mathbf{x}_{\tau}\}_{1 \leq \tau \leq t})$. (b) Prediction network produces output vector $\mathbf{p}_{u} = f^{pred}(\{\mathnormal{y}_{v}\}_{1 \leq v \leq u-1})$ based on the previous non-blank input label. (c) Joint network computes logits by combining the outputs of the previous two networks $z_{t,u} = f^{joint} (\mathbf{h}_{t}^{enc}, \mathbf{p}_{u})$. These logits are then passed to a softmax layer to define a probability distribution. The model can be trained by maximizing the log-likelihood of $P(\mathbf{Y} \in \mathcal{Y}^* | \mathbf{X})$.
\begin{figure}[htb]
	
	\begin{minipage}[b]{.48\linewidth}
		\centering
		\centerline{\includegraphics[width=3.04cm,height=4cm]{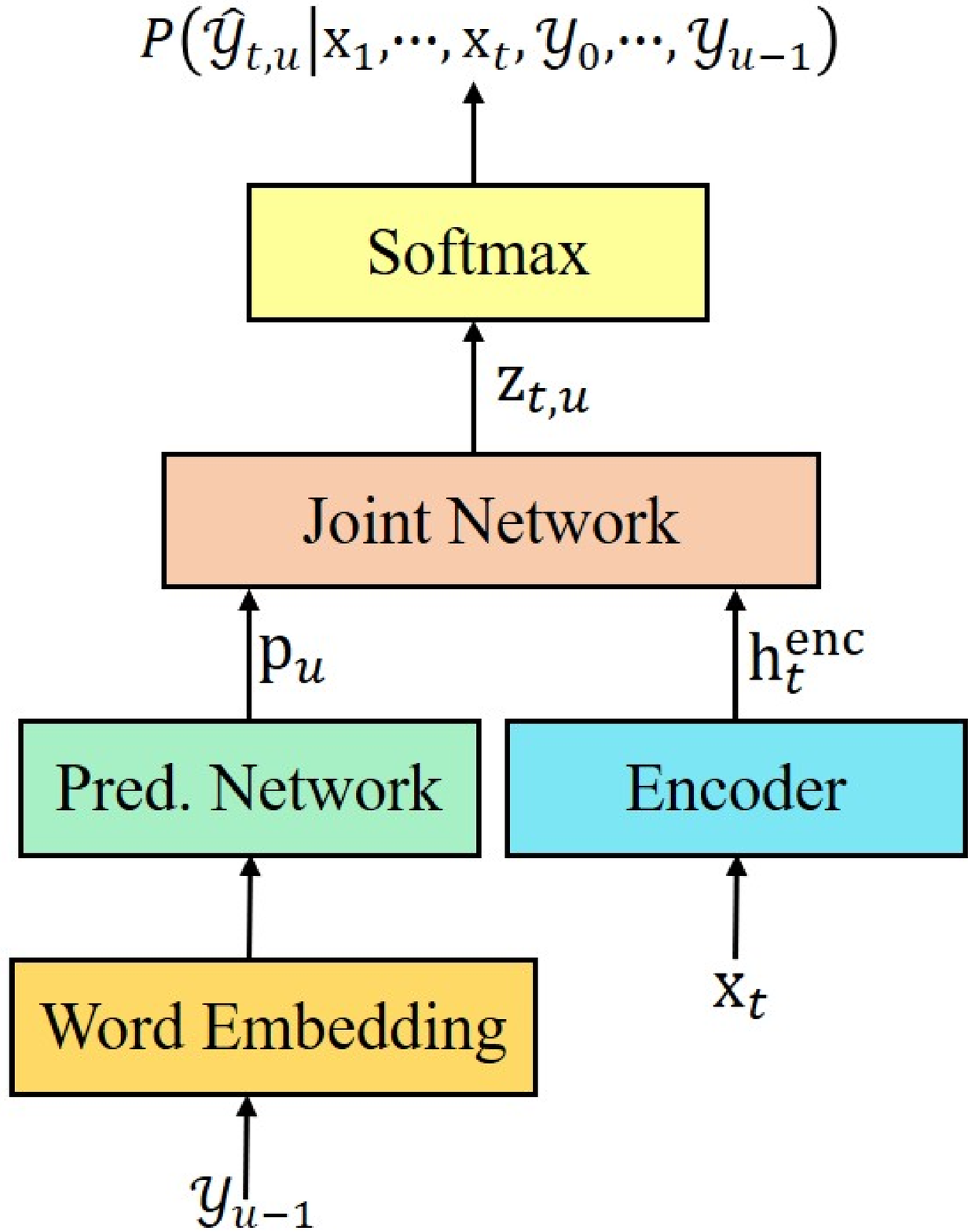}}
		\centerline{(a)}\medskip
	\end{minipage}
	\begin{minipage}[b]{0.48\linewidth}
		\centering
		\centerline{\includegraphics[width=4.5cm,height=4cm]{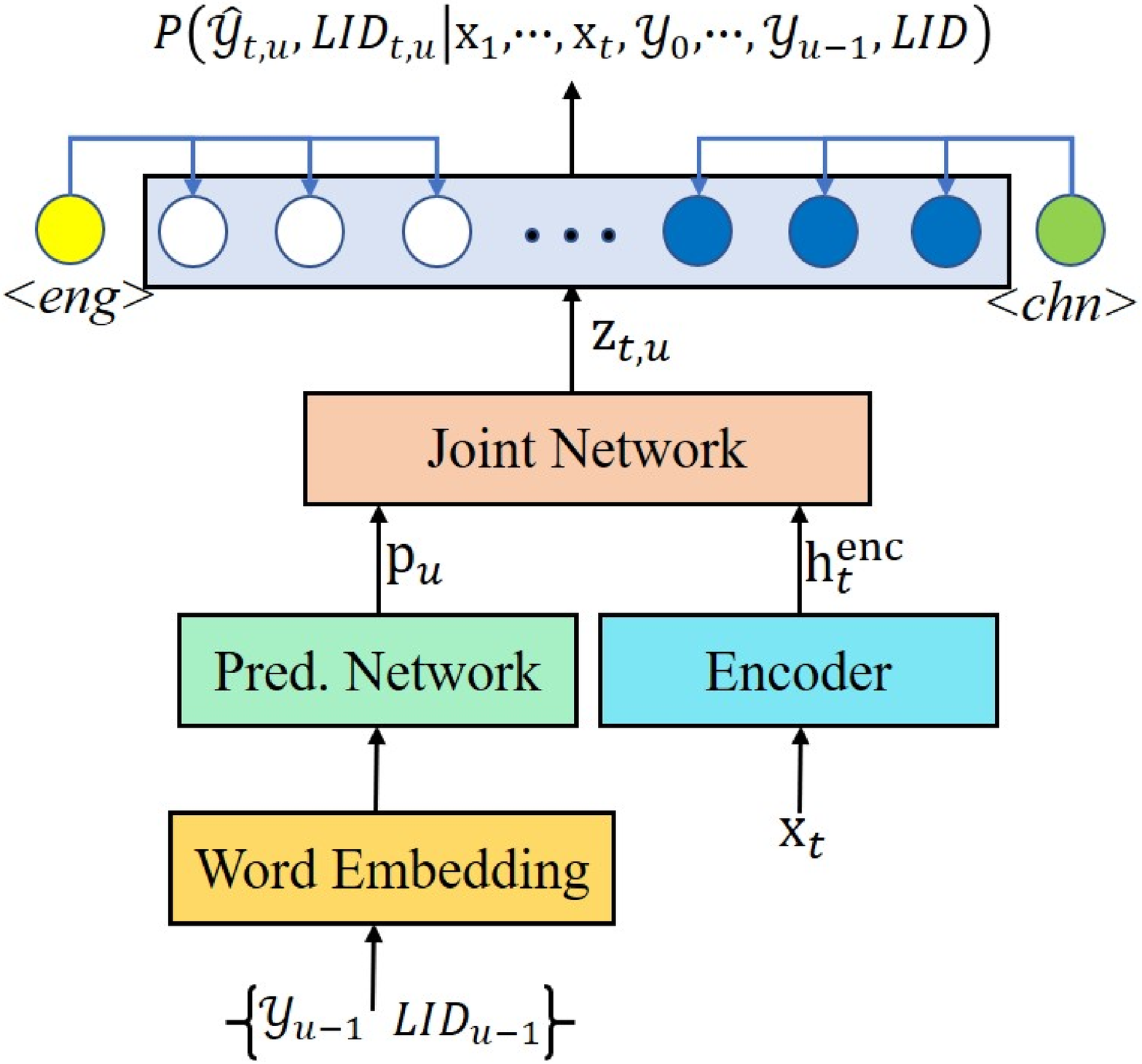}}
		\centerline{(b)}\medskip
	\end{minipage}
	\caption{Basic RNN-T model (a) and RNN-T with language bias (b).}
	\label{fig:res}
\end{figure}

\section{RNN-T with language bias}
\label{sec:pagestyle}

In this paper, we aim to build a concise end-to-end CSSR model that can handle the speech recognition and LID simultaneously. For this task, we augment the output symbols set with language IDs $<chn>$ and $<eng>$ as shown in Fig. 1, i.e., $\hat{\mathcal{Y}} \in \bar{\mathcal{Y}} \cup \{<chn>,<eng>\}$. The intuition behind it is that the CS in the transcript may obey a certain probability distribution, and this distribution can be learned by neural network. 

The properties of RNN-T is key for the problem. It can predict rich set of target symbols such as speaker role and "end-of-word" symbol, which are not related to the input feature directly \cite{he2017streaming,shafey2019joint}. So the language IDs can also be treated as the output symbols. What's more, RNN-T can seamlessly integrate the acoustic and linguistic information. The prediction network of it can be viewed as an RNN language model which predict the current label given history labels \cite{graves2014towards}. So it is effective in incorporating LID into the language model. In general, predicting language IDs only from text data is difficult. However, the joint training mechanism of RNN-T allows it to combine the language and acoustic information to model the CS distribution. Furthermore, the tagged text can bias the RNN-T to predict language IDs which indicates CS points, yet the model trained with normal text can not do this. That is why we choose RNN-T to build the end-to-end CSSR system. 

To promote the model to learn CS distribution more efficient, We concatenate a short vector to all the English word embedding and the English tag $<eng>$ embedding, another different vector for Mandarin, as shown in the bottom of Fig. 2(b). This enhances the dependence of word embedding to its corresponding language ID. In the training process, RNN-T model can learn the distinction information between the two languages easily. The experiment results show that the word embedding constraint is an effective technology. In the inference process, we use the predicted language ID to adjust the output posteriors, as shown in the head of Fig. 2(b). This can bias the model to predict a certain language words more likely in the next-step decode. Overall, our proposed method can handle the speech recognition and LID simultaneously in a simple way, and without increasing additional burden. This study provides new insights into the CS information of text data and its application in end-to-end CSSR system. As a final note, the training and inference algorithms of the proposed model are similar to the standard RNN-T model.

\section{Experiments setups}

\subsection{Dataset}
\label{ssec:subhead}
We conduct experiments on SEAME (South East Asia Mandarin English), a spontaneous conversational bilingual speech corpus \cite{lyu2015mandarin}. Most of the utterances contain both Mandarin and English uttered by interviews and conversations. We use the standard data partitioning rule of previous works which consists of three parts: $train$, $test_{sge}$ and $test_{man}$ (see Table 1) \cite{guo2018study}. $test_{sge}$ is biased to Southeast Asian accent English speech and $test_{man}$ is biased to Mandarin speech. 

Building an end-to-end model requires lots of training data, we apply speech speed perturbation to augment speech data \cite{ko2017study}. By manipulation, we get 3 times the data, with the speed rate of 0.9, 1, and 1.1 of the original speech respectively. We use the augmented data to build our DNN-HMM system and RNN-T system.

\begin{table}[htb] 
	\centering 
	\caption{Data Statistics of SEAME \cite{guo2018study}} \label{tab:aStrangeTable} %
	\begin{tabular}{cccccc} 
		\hline
		\multirow{2}*{Set}& \multirow{2}*{Speakers} & \multirow{2}*{Hours} & \multicolumn{3}{c}{Duration Ratio (\%)}\\
		\cline{4-6}
		~ & ~ & ~ & Man & En & CS\\
		\hline
		$train$ & 134 & 101.13 & 16 & 16 & 68 \\ 
		\hline 
		$test_{man}$ & 10 & 7.49 & 14 & 7 & 79\\
		\hline 
		$test_{sge}$ & 10 & 3.93 & 6 & 41 & 53 \\
		\hline 
	\end{tabular} 
\end{table}

\subsection{DNN-HMM Baseline System}

\label{ssec:subhead}
In addition to the RNN-T baseline system, we also build a conventional DNN-HMM baseline for comparison. The model is based on time delay neural network (TDNN) which trained with lattice-free maximum mutual information (LF-MMI) \cite{povey2016purely}. The TDNN model has 7 hidden layers with 512 units and the input acoustic future is 13-dimensional Mel-frequency cepstrum coefficient (MFCC). For language modeling, we use SRI language modeling toolkit \cite{stolcke2002srilm} to build 4-gram language model with the training transcription. And we construct the lexicon by combining CMU English lexicon  and our Mandarin lexicon. 

\subsection{RNN-T System}
\label{ssec:subhead}
We construct the RNN-T baseline system as described in Section 3.1. The encoder network of RNN-T model consists of 4 layers of 512 long short-term memory (LSTM). The prediction network is 2 layers with 512 LSTM units. And the joint network consists of single feed-forward layer of 512 units with tanh activate function. 

The input acoustic features of encoder network are 80-dimensional log Mel-filterbank with 25ms windowing and 10ms frame shift. Mean and normalization is applied to the futures. And the input words embedding of prediction network is in 512 dimensions continuous numerical vector space. During training, the ADAM algorithm is used as the optimization method, we set the initial learning rate as 0.001 and decrease it linearly when there is no improvement on the validation set. To reduce the over-fitting problem, the dropout rate is set to 0.2 throughout all the experiments. In the inference process, the beam-search algorithm \cite{graves2013speech} with beam size 35 is used to decode the model. All the RNN-T models are trained from scratch use PyTorch.

\subsection{Wordpieces}
\label{ssec:subhead}

For Mandarin-English CSSR task, it is a natural way to construct output units by using characters. However, there are several thousands of Chinese characters and 26 English letters. Meanwhile, the acoustic counterpart of Chinese character is much longer than English letter. So, the character modeling unit will result in significant discrepancy problem between the two languages. To balance the problem, we adopt BPE subword \cite{chiu2018state} as the English modeling units. The targets of our RNN-T baseline system contains 3090 English wordpieces and 3643 Chinese characters. The BPE subword units can not only increase the duration of English modeling units but also maintain a balance unit number of two languages. 

\subsection{Evaluation Metrics}
\label{ssec:subhead}

In this paper, we use mixed error rate (MER) to evaluate the experiment results of our methods. The MER is defined as the combination of word error rate (WER) for English and character error rate (CER) for Mandarin. This metrics can balance the Mandarin and English error rates better compared to the WER or CER.

\section{Results and Analysis}
\label{sec:pagestyle}

\subsection{Results of RNN-T Model}
\label{ssec:subhead}

Table 2 reports our main experiment results of different setups with the standard decode in inference. It is obvious that the MER of end-to-end systems are not as competitive as the LF-MMI TDNN system. The result is consistent with some other reports \cite{zeng2018end}. However, data augmentation is more effective for end-to-end system than TDNN system. It suggests that the gap between our RNN-T and TDNN may further reduce with increasing data. Furthermore, We can also observe that all the experiment results in $test_{sge}$ is much worse than $test_{man}$. This is probably that the accent English in data $test_{sge}$ is more difficult for the recognition system. Bilinguals usually have serious accent problem, which poses challenge to CSSR approaches.

Because the data augmentation technology can significantly reduce the MER of end-to-end model, we conduct all the following experiments based on augmented training data. In order to fairly compare the results of proposed methods with baseline, we remove all the language IDs in the decoded transcription. We can find that The performance of RNN-T model trained (without word embedding constraint) with tagged transcription is much better than the RNN-T baseline. It achieves 9.3\% and 7.6\% relative MER reduction on two test sets respectively. This shows that the tagged text can improve the modeling ability of RNN-T for the CSSR problem. It is the main factor that causes the MER reduction in our experiments. Furthermore, word embedding constraint can also improve the performance of the system though not significant. Overall, our proposed methods yields improved results without increasing additional training or inference burden.

\begin{table}[htb] 
	\centering 
	\caption{The MER of  different setups with standard decode method in inference.} \label{tab:aStrangeTable} %
	\begin{tabular}{cccc} 
		\hline
		\multirow{2}*{Model} & \multirow{2}*{Data Aug}  & \multicolumn{2}{c}{MER(\%)} \\
		\cline{3-4} 
		~ & ~ & $test_{man}$ & $test_{sge}$ \\
		\hline
		\multirow{2}*{TDNN} & No & 24.8 & 33.1  \\ 
		~ & Yes & 23.1 & 31.7 \\
		\hline 
		\multirow{2}*{RNN-T with No-LID} & No & 37.9 & 49.7 \\
		~ & Yes & 33.3 & 44.9 \\
		\hline 
		RNN-T with LID & Yes &  30.2 & 41.5  \\
		++Emb constraint & Yes & 29.5 & 40.3 \\
		\hline 
	\end{tabular} 
\end{table} 

\subsection{Effect of Language IDs Re-weighted Decode}
\label{ssec:subhead}

We then evaluate the system performance by adjusting the weights of next-step predictions in decode process. Table 3 shows the results of RNN-T model with different language IDs weights in inference. It is obvious that the re-weighted methods outperform the model with standard decode process. This suggests that the predicted language IDs can effectively guide the model decoding. 

Because the model assigns language IDs to the recognized words directly, the language IDs error rate is hard to compute. This result may imply that the prediction accuracy of our method is high enough to guide decoding. Meanwhile, We also find that the re-weighted method is more effective on the $test_{man}$ than  $test_{sge}$. This could be caused by higher language IDs prediction accuracy in $test_{man}$. The results of the two different $\lambda$ have similarly MER, and we set $\lambda=0.2$ in the following experiments.
 
\begin{table}[htb] 
	\centering 
	\caption{The MER language IDs re-weighted decode method in inference.} \label{tab:aStrangeTable} %
	\begin{tabular}{ccccc} 
		\hline
		\multirow{2}*{Model}  &  \multirow{2}*{$\lambda$} & \multicolumn{2}{c}{MER(\%)} \\
		\cline{3-4} 
		~  & ~ & $test_{man}$ & $test_{sge}$ \\
		\hline
		\multirow{2}*{RNN-T with LID}  & prob & 29.5 & 41.2\\ 
		~ & 0.2 & 29.4 & 40.9 \\
		\multirow{2}*{++Emb constraint} & prob & 29.0 & 39.9\\
		~ & 0.2 & 28.9 & 39.7 \\
		\hline 
	\end{tabular} 
\end{table}

\subsection{Results of Language Model Re-score}
\label{ssec:subhead}

Table 4 shows the MER results of the N-best (N=35) re-scoring with N-gram and neural language models. The language models are both trained with the tagged training transcription. We see that the language re-scoring can further improve the performance of models. It reveals that the prediction network of RNN-T still has room to be further optimization. Finally, compared to the RNN-T baseline without data augment, the best results of proposed method can achieve 25.9\% and 21.3\% relative MER reduction on two dev sets respectively. compared to the RNN-T baseline with data augment, the proposed method can achieve 16.2\% and 12.9\% relative MER reduction. For both scenarios, our RNN-T methods can achieve better performance than baselines. 

\begin{table}[htb] 
	\centering 
	\caption{The MER of language model rescore.} \label{tab:aStrangeTable} %
	\begin{tabular}{cccc} 
		\hline
		\multirow{2}*{Model} & \multirow{2}*{language model}  & \multicolumn{2}{c}{MER(\%)} \\
		\cline{3-4} 
		~ & ~ & $test_{man}$ & $test_{sge}$ \\
		\hline
		\multirow{2}*{RNN-T with LID} & 4-gram & 28.8  & 40.5\\
		~ & RNN-LM & 28.4  & 39.8\\
		\multirow{2}*{++Emb constraint} & 4-gram  & 28.4 & 39.7\\
		~ & RNN-LM & 27.9 & 39.1 \\
		\hline 
	\end{tabular} 
\end{table}

\section{Conclusions and Future Work}

\label{sec:majhead}
In this work we develop an improved RNN-T model with language bias for end-to-end Mandarin-English CSSR task. Our method can handle the speech recognition and LID simultaneously, no additional LID system is needed. It yields consistent improved results of MER without increasing training or inference burden. Experiment results on SEAME show that proposed approaches significantly reduce the MER of two dev sets from 33.3\% and 44.9\% to 27.9\% and 39.1\% respectively.

In the future, we plan to pre-train the prediction network of RNN-T model using large text corpus, and then finetune the RNN-T model with labeled speech data by frozen the prediction network. 

\section{Acknowledgment}

\label{sec:print}

This work is supported by the National Key Research \& Development Plan of China (No.2018YFB1005003) and the National Natural Science Foundation of China (NSFC) (No.61425017, No.61831022, No.61773379, No.61771472)
\newpage
\bibliographystyle{ieeetran}

\bibliography{refs}

\end{document}